\pdfoutput=1

\documentclass[11pt]{article}

\usepackage[final]{acl}
\usepackage{times}
\usepackage{latexsym}

\usepackage[T1]{fontenc}

\usepackage[utf8]{inputenc}

\usepackage{microtype}

\usepackage{inconsolata}

\usepackage{graphicx}

\usepackage{amsmath}
\usepackage[ruled,vlined]{algorithm2e}
\usepackage{booktabs} 
\usepackage{multirow} 
\usepackage{subcaption}
\usepackage{caption}
\usepackage{xcolor}
\definecolor{softred}{RGB}{190, 40, 40}
\definecolor{softgreen}{RGB}{0, 128, 0}
\title{
Improving Continual Pre-training Through Seamless Data Packing
}

\author{Ruicheng Yin\thanks{\ \ These authors contributed equally.}, Xuan Gao\footnotemark[1], Changze Lv, Xiaohua Wang, \\
{\bf Xiaoqing Zheng\thanks{\ \ Corresponding author.}, Xuanjing Huang} \\
  School of Computer Science, Fudan University, Shanghai, China \\
  \texttt{\{rcyin23,gaox23\}@m.fudan.edu.cn} \\
 \texttt{\{zhengxq,xjhuang\}@fudan.edu.cn} \\}

\begin{document}
\maketitle
\begin{abstract}

Continual pre-training has demonstrated significant potential in enhancing model performance, particularly in domain-specific scenarios.
The most common approach for packing data before continual pre-training involves concatenating input texts and splitting them into fixed-length sequences.
While straightforward and efficient, this method often leads to excessive truncation and context discontinuity, which can hinder model performance. 
To address these issues, we explore the potential of data engineering to enhance continual pre-training, particularly its impact on model performance and efficiency. 
We propose \textbf{Seamless Packing} (SP), a novel data packing strategy aimed at preserving contextual information more effectively and enhancing model performance.
Our approach employs a sliding window technique in the first stage that synchronizes overlapping tokens across consecutive sequences, ensuring better continuity and contextual coherence.
In the second stage, we adopt a First-Fit-Decreasing algorithm to pack shorter texts into bins slightly larger than the target sequence length, thereby minimizing padding and truncation. 
Empirical evaluations across various model architectures and corpus domains demonstrate the effectiveness of our method, outperforming baseline method in 99\% of all settings. 
Code is available at https://github.com/Infernus-WIND/Seamless-Packing.

\end{abstract}

\section{Introduction}

Large Language Models (LLMs) have demonstrated remarkable capabilities in various natural language understanding and generation tasks \cite{NEURIPS2020_1457c0d6, touvron2023llama2openfoundation, yang2024qwen2technicalreport}. These models are usually pre-trained on extensive corpora spanning diverse domains \citep{Radford2018ImprovingLU}, enabling them to perform a wide range of applications. Recent studies have shown that continual pre-training is an effective strategy for adapting LLMs to specific domains or tasks \citep{chen2024towards, jang2022towards, gupta2023continual}, bridging the gap between general-purpose pre-training and task-specific fine-tuning \citep{Biesialska_2020, 10.5555/3648699.3648913}.

Continual pre-training shares many similarities with standard pre-training, particularly in data packing methods, which involve splitting training data into fixed length sequences to facilitate parallelized training.
However, while significant attention has been given to domain-specific data selection, sampling strategies, and mitigating catastrophic forgetting in continual pre-training \citep{lesort2021understanding, winata-etal-2023-overcoming, parmar2024reuse}, relatively little research has focused on the effective organization of training data within batches.

Currently, the most widely adopted method for data packing is straightforward: concatenating all texts into a single stream and segmenting it into fixed-length sequences based on the model’s input size \citep{gururangan-etal-2020-dont}. While efficient and widely used, this approach has a notable drawback—arbitrary truncation of sequences within one piece of text often disrupts contextual continuity, resulting in incomplete sequences. Such disruptions can degrade the model’s ability to understand and generate coherent text, potentially leading to performance issues such as hallucination in downstream tasks \citep{ding2024fewer, achiam2023gpt}. Some existing data packing methods instead rely on padding to ensure uniform sequence lengths. However, padding introduces its own inefficiencies—it occupies valuable input space without any meaningful information, reducing the proportion of real data processed per step. This not only limits training efficiency but may also weaken the model’s ability to leverage longer-range dependencies when a substantial portion of input sequences consists of meaningless tokens.

To address these challenges, we frame the data packing process as an optimization problem aimed at maximizing the contextual continuity within sequences while minimizing truncation and padding, subject to constraints on the pre-defined sequence length and maximum overlapping contexts. 

We propose \textbf{Seamless Packing} (SP) as a solution. 
To optimize sequence utilization, our method follows a two-stage process with a sequential order. 
The first stage prioritizes long texts that meet specific criteria, dynamically generating overlapping sequences while maintaining a predefined maximum repetition ratio to avoid excessive redundancy. 
In conventional data packing, the final portion of a long text often fails to fill a complete sequence, requiring concatenation with subsequent texts. This results in fragmented chunks with insufficient context and additional truncation of the following text. By contrast, the first stage of our method maximizes contextual continuity and minimizes truncation by avoiding such fragmentation, enabling a seamless allocation of sequences, where a long text can fully occupy multiple consecutive sequences without requiring fragments from other texts.

In the second stage, those shorter texts left behind by first stage are packed using an approximation algorithm, First-Fit-Decreasing (FFD). 
The packing bins are assigned a capacity slightly larger than the sequence length to reduce padding, while any tokens exceeding the sequence length are discarded to maintain efficiency. This stage ensures a seamless arrangement of text by minimizing both padding and truncation.
These two stages contribute to seamless text packing from different angles, preserving contextual continuity. By integrating them, our method effectively addresses the optimization problem, resulting in an effective and systematic solution for data packing in continual pre-training.
Furthermore, our method complements and can be integrated with existing approaches that focus on data sampling or mitigating catastrophic forgetting, as it targets a distinct yet important aspect of continual pre-training—optimizing data packing.

To evaluate the effectiveness of our method, we conduct comprehensive experiments on a variety of models and domains. The results demonstrate that our method consistently outperforms conventional data packing methods, achieving superior performance and generalization across diverse settings.
The main contributions are summarized as follows:

\begin{itemize}
\setlength{\itemsep}{0pt}
\setlength{\parsep}{0pt}
\setlength{\parskip}{0pt}
    \item We propose \textbf{Seamless Packing}, a simple yet effective data packing method tailored for continual pre-training. By optimizing both segment placement and overlap strategy, our approach preserves \textit{contextual continuity} while reducing truncation and padding.
    
    \item We present a theoretical analysis of Seamless Packing, characterizing the influence range of overlapping contexts under the control of two key hyperparameters.

    \item We conduct extensive experiments across diverse domains and tasks to validate the effectiveness and generalizability of our method. Results show consistent improvements across domain-specific and general-domain settings.

    \item Through ablation studies, controlled comparisons, we validate the contribution of each component of our method. We also explore the trade-offs between token dropping and padding, and demonstrate how Seamless Packing mitigates truncation-induced hallucination by preserving critical contextual information through a targeted case study.
\end{itemize}

\section{Related Work}

\textbf{Continual Pre-training}\quad 
Continual pre-training has been shown to effectively enhance language models by adapting them to specific domains or improving their general capabilities \citep{gururangan-etal-2020-dont, yildiz2024investigating}. This approach involves a second phase of pre-training on unlabeled data, allowing models to integrate new knowledge without requiring extensive labeled datasets. 

\begin{figure*}[t] 
    \centering
    \includegraphics[width=0.85\textwidth]{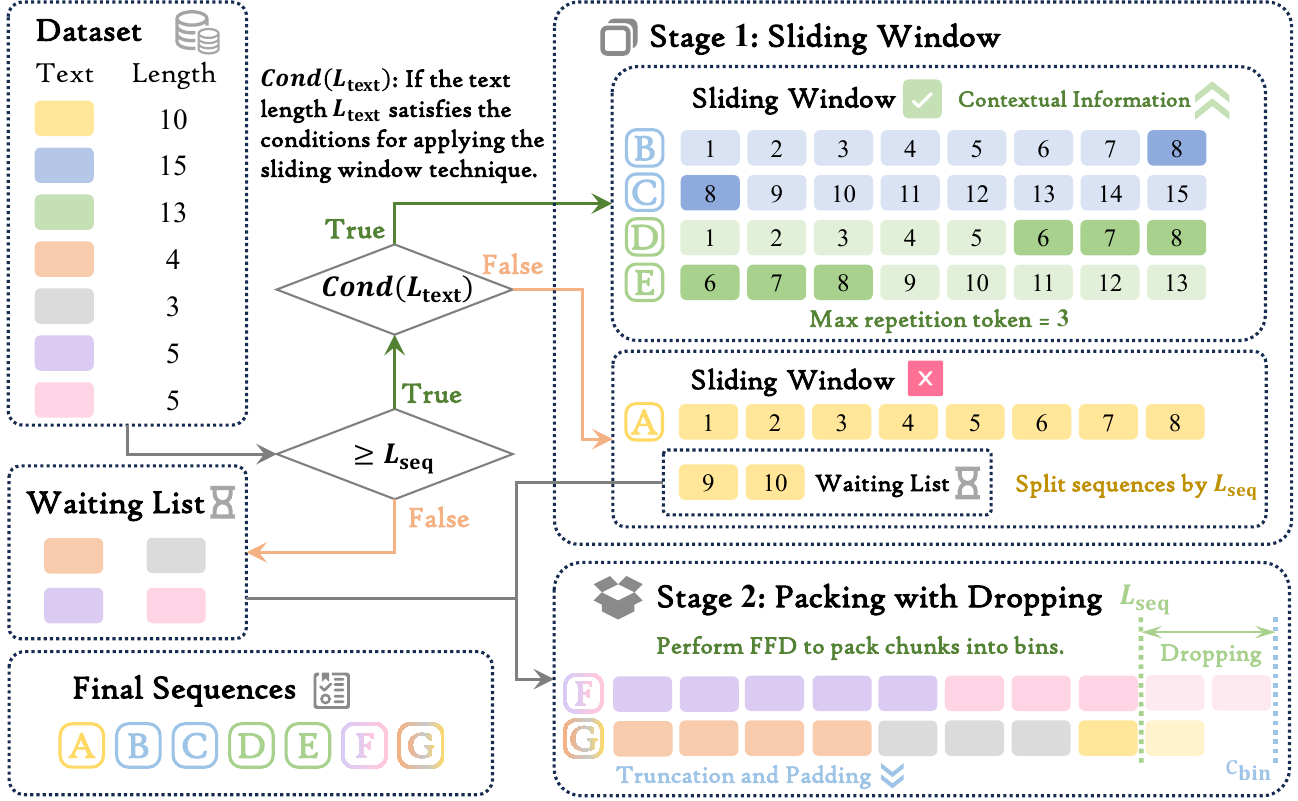}
    \caption{An illustration of the proposed Seamless Packing method. \textbf{Top-right}: Stage 1, each text is assessed to determine whether the sliding window technique can be applied. \textbf{Top-right (upper)}: Sliding window applicable—darker-colored tokens indicate repeated tokens. \textbf{Top-right (lower)}: Sliding window not applicable—chunks that cannot fill an entire sequence are added to the waiting list. \textbf{Bottom-right}: Stage 2, short chunks are packed using the First-Fit-Decreasing strategy. Lighter-colored tokens represent discarded tokens. Note: $L_\text{seq}$ -  sequence length, $c_\text{bin}$ - bin capacity. Both $L_\text{seq}$ and $c_\text{bin}$ are hyperparameters, which are set to $L_\text{seq} = 8$ and $c_\text{bin} = 10$ in this figure.}
    \label{fig:mainfigure}
\end{figure*}

\noindent \textbf{Data Packing}\quad 
Several studies have explored data packing strategies to optimize language model training. \citet{krell2021efficient} proposed a packing algorithm designed to maximize sequence length utilization, thereby reducing the need for padding tokens. \citet{ding2024fewer} conducted both theoretical and empirical analyses on the impact of truncation, emphasizing that preserving data integrity is crucial for improving language modeling performance. To address this, they introduced a Best-Fit-Decreasing packing algorithm based on segment trees, which optimally reduces truncation. 
Our proposed method integrates the strengths of these approaches while introducing novel innovations, achieving a balanced optimization of the data packing process that preserves contextual continuity and minimizes truncation and padding, leading to a seamless packing algorithm.
More information is provided in Appendix~\ref{sec:appendix_relatedwork}.

\section{Method}
In this section, we introduce Seamless Packing in details.
As illustrated in Figure~\ref{fig:mainfigure}, our method consists of two stages: Sliding Window and Packing with Dropping. 
The design of the two stages is motivated by the need to address two crucial challenges in data packing for continual pre-training: preserving contextual continuity and minimizing truncation and padding, the necessity of which has been highlighted in previous section. 
In the first stage, long texts that meet certain criteria are processed using a dynamic sliding window technique, effectively leveraging contextual continuity and minimizing truncation. 
The second stage processed the remaining short texts, reducing both truncation and padding. 
Both stages are carefully designed to ensure the algorithm seamless and improve the overall performance of LLMs in continual pre-training scenarios.

\subsection{The Importance of Contextual Continuity}

Before introducing our packing strategy, we first examine how naive truncation—commonly used in continual pre-training pipelines—can disrupt contextual continuity. While efficient, this preprocessing approach often fragments semantically connected content, undermining the model’s ability to reason over longer contexts.

First, consider a sentence like \textit{``Event A is [DESCRIPTION], which will be held on [DATE], in [PLACE].''} If the input is segmented just before ``which'', the event description remains in one chunk, while time and location details appear in another. This disconnection can hinder the model’s ability to associate the components of a single event.

Second, in summarization tasks, truncating key explanatory context may result in unfaithful outputs. For example, if \textit{``The Supreme Court ruled in favor of the plaintiff after considering key arguments related to privacy law''} is split before ``after considering...,'' the generated summary may omit or hallucinate the basis for the ruling.

Third, in code understanding, a variable declaration split from its usage—e.g., when the definition appears in one segment and its reference in another—can lead to incomplete or incorrect completions, as dependencies are no longer visible \citep{ding2024fewer}.

These examples underscore the importance of preserving contextual continuity in data preprocessing. To address this, our Seamless Packing method is designed to minimize context fragmentation through overlapping windows and adaptive sequence packing. In Section~\ref{subsec:CaseStudy}, we further support this analysis with a case study that empirically demonstrates the detrimental effects of truncation.

\subsection{First Stage: Sliding Window}
The first stage of our method is inspired by the sliding window technique  \citep{10.1145/359842.359859, karp1987efficient, KOC199517}, which is widely employed in algorithm design across various fields. Traditional sliding window uses a fixed stride to determine how much the window advances at each step. However, a fixed stride is suboptimal in Seamless Packing. \textit{Fixed} strides cannot dynamically adapt to varying texts lengths, leading to either incomplete coverage or excessive overlap, which disrupts seamless alignment of sequences. Moreover, an overly small stride results in a high overlap ratio between consecutive sequences, reducing training efficiency and potentially degrading model performance, as observed in prior studies \citep{10.1145/3359591.3359735, lee-etal-2022-deduplicating}. Conversely, larger stride values complicate the intuitive understanding of the sliding window’s purpose in our method, which is to facilitate context sharing across sequences. 

To overcome these limitations, we introduce a more intuitive parameter, maximum repetition ratio, denoted as \( r_{\text{max}} \), which offers a more precise control over sequence overlap, ensuring optimal training dynamics and model efficacy. Given a tokenized text of length \( L_{\text{original}} \) that can be divided into \( n \) full sequences of length \( L_{\text{seq}} \), the maximum number of overlapping tokens allowed is: 
\begin{equation}
\small
L_{\text{max\_overlap}} = \lceil n \times r_{\text{max}} \times L_{\text{seq}} \rceil.
\tag{1} 
\end{equation}
The sliding window technique is applicable when:  
\begin{equation}
\small
L_{\text{original}} + L_{\text{max\_overlap}} \geq (n+1) \times L_{\text{seq}},
\tag{2}
\end{equation}
which ensures that the text can be expanded to fill \( n+1 \) sequences. In such cases, the overlap is dynamically adjusted to achieve full sequence utilization: 
\begin{equation}
\small
L_{\text{final\_overlap}} = \lceil \frac{(n+1) \times L_{\text{seq}} - L_{\text{original}}}{n} \rceil.
\tag{3}
\end{equation} 
This strategy ensures an efficient balance between overlap and utilization of available sequence space.  

For texts that do not satisfy the condition described in expression (2), we simply divide them into chunks of length \( L_{\text{seq}} \). Chunks that entirely occupy a sequence are retained, while incomplete chunks are deferred to the second stage for further processing.

The advantages of our sliding window technique are twofold. First, it allows some sequences to incorporate additional context compared to the traditional method. Second, it reduces cases where chunks from different documents are concatenated within a single sequence. While such cases are handled by the model through adding special token or 
attention masking \citep{NIPS2017_3f5ee243, devlin-etal-2019-bert}, the contextual information of chunks in such situation is often limited, offering minimal benefit to training. By mitigating these issues, our method enhances context preservation and ultimately improves model performance.

\subsection{Second Stage: Packing with Dropping}
The first stage has processed all chunks that can occupy an entire sequence, leaving short chunks behind. Thus, the main purpose of second stage is to pack these short chunks together efficiently to maximize sequence utilization.

At its core, this stage is formulated as a variant of the bin packing problem \citep{MARTELLO199059}, which seeks to minimize the number of fixed-capacity bins required to accommodate items of varying sizes. As this problem is NP-hard, we employ an approximation algorithm, specifically the First-Fit-Decreasing (FFD) \citep{johnson1973near} heuristic. While Best-Fit-Decreasing (BFD) \citep{eilon1971loading} is another widely used heuristic, we adopt FFD because its early stopping mechanism significantly reduces computation time while achieving competitive performance, as we will show in Section \ref{subsec:ablation}.

\begin{algorithm}[!tb]
\caption{First-Fit-Decreasing}
\label{alg:ffd}
\KwIn{Texts $T = \{t_i\}_{i=1}^N$}
\KwOut{Bins $\{b_j\}$ containing all items}
Define $l(t)$: length of text $t$\;
Define $rc(b)$: remaining capacity of bin $b$\;
Sort $T$ in descending order of length, resulting in $T_s$\;
Initialize an empty set of bins: $\{b_i\}_{i=1}^N$\;
\ForEach{$t_i \in T_s$}{
    Find the first bin $b_j$ such that $rc(b_j) \geq l(t_i)$\;
    Add $t_i$ to $b_j$\;
}
\end{algorithm}

While the FFD algorithm we employ follows a standard implementation, we introduce a flexible modification to better adapt to different dataset characteristics. Specifically, we allow the bin capacity to be slightly larger than \( L_{\text{seq}} \), controlled by a parameter called extra bin capacity, denoted as \( c_{\text{extra}} \), and tokens that exceed \( L_{\text{seq}} \) after packing are discarded. This adjustment provides an alternative strategy rather than strictly relying on padding, making the approach adaptable to different scenarios. While padding could be used as an alternative under certain conditions, we firmly adopt the dropping strategy in this study and demonstrate its effectiveness across all experimental settings. Our results show that dropping consistently achieves competitive performance, reinforcing its suitability as the preferred choice.

The rationale behind this design is twofold. First, dropping eliminates the inefficiencies associated with excessive padding, ensuring that training is not diluted by meaningless tokens and maintains a seamless algorithm. Second, the number of discarded tokens can be precisely controlled by \( c_{\text{extra}} \), and when properly tuned, the fraction of dropped tokens remains negligible, having minimal impact on overall performance. The complete FFD algorithm we used is outlined in Algorithm \ref{alg:ffd}.

After applying the above procedure, a small number of bins may still contain sequences whose total length shorter than \( L_{\text{seq}} \). At this point, the remaining data volume is minimal. To ensure that Seamless Packing remains a padding-free method, we concatenate all remaining chunks into a single sequence, which is then divided into chunks of length \( L_{\text{seq}} \) and added to the final output.

The primary objective of the Packing with Dropping approach is to process shorter text chunks while avoiding unnecessary truncation or padding, both of which have been previously identified as limitations. By minimizing these inefficiencies, this stage further contributes to improving the overall performance of the model.

\subsection{Theoretical Analysis}
\label{subsec:TheoreticalAnalysis}
We now conduct a theoretical analysis of the proportion of texts that satisfy the condition in expression (2) and the amount of shorter chunks processed in stage two, as influenced by \( r_{\text{max}} \). This analysis allows us to understand the proportion of data affected by each stage of our method, providing a more comprehensive view of its impact.

Before delving into the analysis, we first examine the dataset's length distribution. We present the token count distribution of datasets we used in Appendix~\ref{sec:dataset_length_distribution}. A key observation is that the number of texts within each length interval decreases as the sequence length increases. While the actual distribution is unlikely to follow a perfectly linear trend, the samples within each interval can be reasonably approximated as uniformly distributed for analytical purposes.

By combining expressions (1) and (2), we derive the condition for applying the sliding window:  
\begin{equation}
\small
L_{\text{original}} \geq (n+1 - n r) \times L_{\text{seq}}.
\tag{4}
\end{equation}
Given the definition of \( n \), we also have:  
\begin{equation}
\small
L_{\text{original}} \geq n \times L_{\text{seq}}.
\tag{5}
\end{equation}
Since expression (5) always holds, if \( n+1 - n r \leq n \) is satisfied, expression (4) follows directly. This implies that when \( n \geq \lceil \frac{1}{r} \rceil \), all relevant texts can be processed using the sliding window.  

For intervals where \( n < \lceil \frac{1}{r} \rceil \), the proportion of text samples within each interval that can be processed using the sliding window is \( n r \). Therefore, the total number of texts \(N_{sw}\) that can be processed using the sliding window is given by:  
\begin{equation}
\small
N_{sw} = \sum_{k=1}^{\lceil \frac{1}{r} \rceil} (k r T_k) + \sum_{k=\lceil \frac{1}{r} \rceil}^{M} T_k,
\tag{6}
\end{equation}
which evidently increases as \( r \) increases and \( T_k \) denotes the number of texts whose tokenized length falls within the interval \( (kL_{\text{seq}}, (k+1)L_{\text{seq}}] \), where \( k \in \{1, 2, \dots, M\} \) and \( M \) represents the largest interval considered in the dataset.

\begin{table*}[t]
\centering
\resizebox{\textwidth}{!}{
\begin{tabular}{lcccccccccccc}
\toprule
\multirow{2}{*}{\textbf{Tasks}} & \multicolumn{4}{c}{\textbf{GPT2 - large (812M)}} & \multicolumn{4}{c}{\textbf{Llama3.2 - 1B}} & \multicolumn{4}{c}{\textbf{Qwen2.5 - 1.5B}} \\ 
\cmidrule(lr){2-5} \cmidrule(lr){6-9} \cmidrule(lr){10-13}
& OM & CT & BFD & SP & OM & CT & BFD & SP & OM & CT & BFD & SP \\ 
\midrule
\midrule
BBC News & 97.18 & 96.80 & 97.22 & \textbf{97.38} & 97.22 & 97.44 & 97.54 & \textbf{97.64} & 97.68 & 97.74 & 97.80 & \textbf{98.06} \\
AG News & 89.29 & 88.66 & 88.32 & \textbf{89.65} & 89.76 & 89.58 & 89.97& \textbf{90.00} & 88.53 & 89.39 & 89.71 & \textbf{90.37} \\
20 Newsgroup & 65.76 & 66.35 & 65.60 & \textbf{67.13} & 65.73 & 64.74 & 65.71 & \textbf{66.30} & 67.11 & 67.78 & 67.24 & \textbf{68.02} \\
\midrule
Fin Topic & 81.39 & 81.71 & 81.79 & \textbf{81.99} & 82.84 & 82.69 & \textbf{84.62} & 84.58 & 81.99 & 83.13 & 82.69 & \textbf{84.18} \\
Fin Sentiment & 85.33 & 86.79 & 87.08 & \textbf{87.29} & 87.25 & 88.18 & 88.18 & \textbf{88.68} & 86.92 & 88.51 & 88.68 & \textbf{88.85} \\
Fin Phrasebank & 82.78 & 82.10 & 82.94 & \textbf{83.04} & 82.06 & 83.96 & 83.70 & \textbf{84.44} & 84.08 & 84.02 & 84.12 & \textbf{84.52} \\
\midrule
PubMed Class & 84.62 & 84.69 & 85.17 & \textbf{85.88} & 83.09 & 84.18 & 83.67 & \textbf{84.79} & 85.10 & 86.32 & 86.22 & \textbf{86.87} \\
ChemProt & 80.96 & 81.48 & 81.51 & \textbf{82.35} & 77.71 & 80.03 & \textbf{80.81} & \textbf{80.81} & 81.39 & 81.89 & 81.83 & \textbf{82.64} \\
\bottomrule
\end{tabular}}
\caption{Performance on downstream tasks after full parameter fine-tuning, comparing original pre-trained model and three different data packing methods used in continual pre-training.  Note: OM - Original Model, CT - Concatenation and Truncation, BFD - Best Fit Decreasing, SP - Seamless Packing, Fin - Financial, class - classification}
\label{tab:maintableperformance}
\end{table*}

Next, we consider the texts that cannot be processed using the sliding window, which are divided into shorter chunks. The total number of tokens for these short texts is given by:
\begin{equation}
\small
N_{\text{token\_short}} = \sum_{k=1}^{\lceil \frac{1}{r} \rceil} \left[(1 - k r) T_k \times \frac{(1 - k r) L_{\text{seq}}}{2}\right].
\tag{7}
\end{equation}
In this expression, the first part, \( (1 - k r) T_k \), represents the number of short texts within each interval that cannot be processed using the sliding window. The second part, \( \frac{(1 - k r) L_{\text{seq}}}{2} \), is the average length of the short texts in each interval, assuming a uniform distribution of lengths within the interval. Thus, the formula computes the total number of tokens in all these short texts. 
We further illustrate our theoretical analysis in appendix~\ref{sec:appendix_theoretical_analysis}.

\section{Experiments}
We conduct extensive experiments to assess the effectiveness of Seamless Packing across diverse domains and models. Unless otherwise specified, models are independently pre-trained on domain-specific corpora. Additionally, we design ablation studies to validate the contributions of both stages in our method. Through comprehensive evaluations, we analyze the impact of hyperparameters, validate the theoretical analysis and further investigate key aspects of our method. The results demonstrate that our method consistently outperforms traditional approaches across all tasks, achieving a fourfold improvement over baseline methods (0.96\% vs. 0.24\%). 
All fine-tuning results are averaged over five runs (seeds 42–46).

\subsection{Datasets and Baselines}
The dataset used for continual pre-training are BBC news \citep{li2024latesteval}, financial news article \citep{financearticle} and academic articles from PubMed \citep{cohan-etal-2018-discourse}. Our experiments are conducted on GPT-2 \citep{radford2019language}, LLaMA-3 \citep{dubey2024llama}, Qwen2.5 \citep{qwen2.5} and Gemma-2 \citep{team2024gemma}. The universal baselines consist of three configurations: the original model, the model continual pre-trained through traditional concatenation and truncation, and the model continual pre-trained using BFD algorithm. 
More details are presented in Appendix~\ref{sec:appendix_experiment_related}.

\subsection{Main Results}
\label{subsec:performance}

\begin{table}[t]
\centering
\scalebox{0.8}{
\begin{tabular}{ccccc}
\toprule
\textbf{Model} & \textbf{Domain} & \textbf{CT} & \textbf{BFD} & \textbf{SP} \\ 
\midrule
\midrule
\multirow{3}{*}{\shortstack{GPT2 \\ \\ 812M}} 
& News & 10.79 & 10.28 & \textbf{10.13} \\ 
& Finance & 7.19 & 6.74 & \textbf{6.73} \\
& Med & 11.49 & 10.16 & \textbf{10.12} \\ 
\midrule
\multirow{3}{*}{\shortstack{Llama3.2 \\ \\ 1B}} 
& News & 11.48 & 10.52 & \textbf{10.07}  \\
& Finance & 6.27 & 5.71 & \textbf{5.52}  \\ 
& Med & 7.84 & 7.41 & \textbf{7.39}  \\ 
\midrule
\multirow{3}{*}{\shortstack{Qwen2.5 \\ \\ 1.5B}} 
& News & 9.51 & 9.20 & \textbf{9.19} \\
& Finance & 4.76 & 4.56 & \textbf{4.50} \\
& Med & 6.71 & 6.47 & \textbf{6.43} \\ 
\bottomrule
\end{tabular}}
\caption{Perplexity on the validation dataset during continual pre-training, comparing three different data packing methods on three models and three domains respectively.}
\label{tab:perplexity}
\end{table}

\textbf{Continual pre-training perplexity}\quad
Perplexity is a widely used evaluation metric in natural language processing to measure the quality of a language model's predictions \citep{jelinek1977perplexity, miaschi-etal-2021-makes}. In the context of continual pre-training, perplexity serves as a key indicator of the effectiveness of the training approach. As shown in Table~\ref{tab:perplexity}, our proposed method achieves a significantly lower perplexity on the validation set compared to the baseline approach, demonstrating its superior ability to capture linguistic structure and generalize to unseen data. This improvement underscores the effectiveness of Seamless Packing in leveraging overlapping contexts to enhance the model's understanding and predictive accuracy.

\noindent \textbf{Full parameter fine-tuning}\quad
We evaluate the performance of our method on eight diverse downstream tasks spanning the three selected domains. The models are fine-tuned on downstream tasks after continual pre-training using different methods. The results are presented in Table~\ref{tab:maintableperformance} and statistical significance tests are summarized in Appendix~\ref{sec:appendix_stats_significance}.

The results show that while performance gains vary across models and tasks, our method consistently achieves the best results in the majority of cases. This highlights not only the strong efficacy of our approach but also its robustness across different domains and task types with diverse linguistic complexity and data distributions.

Additionally, our method exhibits stable gains in scenarios where baseline methods struggle. 
For every model, BFD baseline method under-performs traditional method of concatenation and truncation in 2 to 4 tasks, whereas our method maintains robust performance across all tasks without such failures. This suggests that our improvements are both significant and broadly applicable across different task characteristics.

Overall, the empirical results support our hypothesis that our method optimizes contextual understanding during continual pre-training, leading to better performance. The consistent improvements across domains further confirm its scalability.

\begin{table}[t]
\centering
\scalebox{0.8}{
\begin{tabular}{ccccc}
\toprule
\textbf{Model} & \textbf{Metric/Task} & \textbf{CT} & \textbf{BFD} & \textbf{SP} \\ 
\midrule
\midrule
\multirow{4}{*}{\shortstack{Qwen2.5 \\ \\ 3B}} 
& Perplexity & 10.27 & 9.40 & \textbf{9.24} \\ 
\cmidrule(lr){2-5}
& BBC News & 97.54 & \textbf{97.56} & 97.50 \\ 
& AG News & 88.13 & 87.21 & \textbf{88.66}  \\ 
& 20 Newsgroup & 58.65 & 58.95 & \textbf{59.38} \\ 
\midrule
\multirow{3}{*}{\shortstack{Gemma2 \\ \\ 2B}} 
& Perplexity & 11.11 & \textbf{10.97} & 11.02 \\ 
\cmidrule(lr){2-5}
& PubMed Class & 85.03 & 84.86 & \textbf{85.10} \\ 
& ChemProt & 65.98 & 63.95 & \textbf{66.47}  \\ 
\bottomrule
\end{tabular}}
\caption{Perplexity on the validation dataset during continual pre-training and performance on downstream tasks after LoRA tuning, comparing three different data packing methods.}
\label{tab:lora}
\end{table}

\noindent \textbf{LoRA tuning}\quad
LoRA (Low-Rank Adaptation, \citealp{hu2022lora}) is a widely adopted parameter-efficient fine-tuning (PEFT) method that updates only a small subset of parameters, making it particularly effective for adapting large pre-trained models to specific tasks with minimal computational overhead. We include this set of experiments to demonstrate that our method not only excels under full-parameter fine-tuning but also achieves strong performance with PEFT approaches such as LoRA. As shown in Table~\ref{tab:lora}, our method consistently achieves competitive results under LoRA tuning, further highlighting its adaptability to mainstream fine-tuning paradigms.

\subsection{Generalization Analysis}
\label{sec:generalization-analysis}
To assess the robustness and adaptability of Seamless Packing beyond specific data configurations, we conduct experiments under two distinct settings: mixed-domain and general-domain.

\noindent \textbf{Mixed-Domain}\quad
We pre-train a model on a mixed-domain corpus that combines three specialized domains: News, Finance, and Medical. Evaluation is conducted on three downstream tasks, each corresponding to one of the original domains. This setup enables us to test the model's ability to retain and integrate heterogeneous domain-specific knowledge. Table~\ref{tab:mixed-domain} reports the performance comparison among different packing strategies.

\begin{table}[t]
\centering
\scalebox{0.8}{
\begin{tabular}{lcccc}
\toprule
\textbf{Model} & \textbf{Task} & \textbf{CT} & \textbf{BFD} & \textbf{SP} \\
\midrule
\midrule
\multirow{3}{*}{\shortstack{Llama3.2 \\ \\ 1B}} & 20 Newsgroup   & 68.72 & 68.55 & \textbf{69.80} \\
& Fin Phrasebank & 84.10 & 84.04 & \textbf{84.58} \\
& ChemProt       & 81.22 & 82.14 & \textbf{82.58} \\
\bottomrule
\end{tabular}}
\caption{Evaluation results of mixed-domain continual pre-training.}
\label{tab:mixed-domain}
\end{table}

\noindent \textbf{General-Domain}\quad
To examine whether SP generalizes beyond domain-specific settings, we adopt RedPajama \citep{weber2024redpajama}, an open-source dataset designed to approximate the original LLaMA training corpus. To avoid overlap with the original LLaMA training data, we select GPT-2-Large as the base model and conduct continual pre-training on a sampled subset of RedPajama. The model is evaluated on two domain-specific downstream tasks (20 Newsgroup, ChemProt) to assess cross-domain transferability, and on three general-purpose tasks from the GLUE benchmark (\citet{wang-etal-2018-glue}, MNLI, QNLI, RTE) to evaluate natural language understanding capabilities. Results are shown in Table~\ref{tab:general}.

\begin{table}[t]
\centering
\scalebox{0.8}{
\begin{tabular}{lcccc}
\toprule
\textbf{Model} & \textbf{Task} & \textbf{CT} & \textbf{BFD} & \textbf{SP} \\
\midrule
\midrule
\multirow{5}{*}{\shortstack{GPT2 \\ \\ 812M}} & 20 Newsgroup & 69.15 & 69.42 & \textbf{69.69} \\
& ChemProt     & 81.62 & 80.64 & \textbf{82.50} \\
\cmidrule(lr){2-5}
& MNLI         & 68.95 & 67.93 & \textbf{70.47} \\
& QNLI         & 80.77 & 80.40 & \textbf{81.80} \\
& RTE          & 69.53 & 69.31 & \textbf{71.33} \\
\bottomrule
\end{tabular}}
\caption{Evaluation results of general-domain continual pre-training.}
\label{tab:general}
\end{table}

\noindent \textbf{Discussion}\quad
Across both experiments, SP consistently achieves superior performance. Whether operating in mixed-domain settings or pre-training from general-domain corpora, our method outperforms both naive concatenation and standard bin-packing strategies. These findings confirm that SP generalizes effectively across both domain-specific and general-domain continual pre-training, supporting its applicability in a wide range of real-world scenarios.

\begin{table}[t]
\centering
\resizebox{\columnwidth}{!}{
\begin{tabular}{lccccc}
\toprule
\textbf{Model} & \textbf{Task} & \textbf{OM} &\textbf{CT} & \textbf{BFD} & \textbf{SP} \\
\midrule
\midrule
Llama3.2 1B & XNLI (French) & 63.87 & 67.72 & 68.64 & \textbf{69.04} \\
\bottomrule
\end{tabular}}
\caption{Evaluation results of the French XNLI task.}
\label{tab:french}
\end{table}

\subsection{Cross-lingual Transfer}
To further evaluate the cross-lingual applicability of Seamless Packing, we conduct experiments on a French dataset. Specifically, we continually pre-train Llama-3.2-1B on a sampled French subset of the C4 corpus \citep{raffel2020exploring} and evaluate the model on the French NLI task from the FLUE benchmark (\citet{le2019flaubert}, French Language Understanding Evaluation). The XNLI dataset used here provides a direct measure of a model's ability to perform natural language inference in a non-English language.
Table~\ref{tab:french} presents the result and our method again yields the highest performance.

These results demonstrate that Seamless Packing is effective even in cross-lingual scenarios, highlighting its potential to support multilingual continual pre-training and downstream task adaptation.

\begin{table}[t]
\centering
\resizebox{\columnwidth}{!}{
\begin{tabular}{cccccc}
\toprule
\textbf{Metric/Task} & \textbf{Model} & \textbf{FFD} & \textbf{BFD} & \textbf{BFD-m} & \textbf{SP} \\ 
\midrule
\midrule
\multirow{3}{*}{\shortstack{Perplexity}} 
& Qwen2.5 0.5B & \textbf{11.24} & \textbf{11.24} & \textbf{11.24} & \textbf{11.24} \\ 
& Llama3.2 1B & 12.08 & 12.39 & \textbf{11.95} & \underline{11.99} \\ 
& Qwen2.5 1.5B & 9.45 & \textbf{9.39} & \underline{9.40} & 9.48 \\
\midrule
\multirow{3}{*}{\shortstack{PubMed Class}} 
& Qwen2.5 0.5B & 85.17 & 85.10 & \underline{85.23} & \textbf{86.12} \\ 
& Llama3.2 1B & 85.00 & 84.89 & \underline{85.37} & \textbf{85.47} \\ 
& Qwen2.5 1.5B & 85.78 & 85.85 & \underline{86.11} & \textbf{86.56} \\ 
\midrule
\multirow{3}{*}{\shortstack{ChemProt}} 
& Qwen2.5 0.5B & 78.78 & 78.64 & \underline{79.07} & \textbf{80.26} \\ 
& Llama3.2 1B & 80.38 & 80.38 & \underline{80.44} & \textbf{81.45} \\ 
& Qwen2.5 1.5B & 81.83 & 82.00 & \underline{82.09} & \textbf{82.53} \\ 
\bottomrule
\end{tabular}}
\caption{Perplexity on the validation dataset during continual pre-training and performance on downstream tasks after fine-tuning, comparing four different data packing methods. The underlined text represents the second-best performance.}
\label{tab:ablation}
\end{table}

\subsection{Ablation Study}
\label{subsec:ablation}
We conduct a series of ablation studies to evaluate the contributions of the two key stages in Seamless Packing and to compare BFD and FFD. We select three different models and continually pre-train them on a resampled PubMed article dataset using various data packing methods. We verify model's performance on two downstream tasks with full parameter fine-tuning. Additionally, we introduce two alternative methods: FFD and BFD-m. FFD employs the FFD algorithm to pack all texts into bins with a capacity of \( L_{\text{seq}} \), whereas BFD-m extends BFD by allowing a larger bin capacity, exceeding \( L_{\text{seq}} \). BFD-m serves as an intermediate variant to isolate effects of two stages in Seamless Packing: comparing BFD with BFD-m reveals contribution of the second stage, while comparing BFD-m with SP highlights impact of the first stage.

As shown in Table \ref{tab:ablation}, although Seamless Packing does not always achieve the lowest perplexity, it consistently delivers the best downstream task performance in most cases, outperforming BFD-m. This result highlights the effectiveness of the sliding window technique in our approach. Furthermore, BFD-m outperforms standard BFD in all scenarios, suggesting that slightly increasing bin capacity enhances overall model performance.

Interestingly, FFD performs comparably to BFD on average, indicating that BFD may be unnecessary in scenarios where it does not provide a clear performance advantage but incurs higher computational costs. A detailed time consumption analysis is provided in Section~\ref{subsec:EmpiricalAnalysis}.

We also conducted experiments to evaluate the impact of optimizer and scheduler choices. Detailed results are presented in Appendix~\ref{sec:appendix_lr}.

\subsection{Hyperparameters}
Seamless Packing involves two key hyperparameters:  \( r_{\text{max}} \) and \( c_{\text{extra}} \). This subsection analyzes their effects and presents experimental results to determine optimal settings. The experiment setup follows the same configuration as in Section \ref{subsec:ablation}. The complete results are presented in Appendix~\ref{sec:appendix_hyperparameter}.

\begin{figure}[t]
    \centering
    \begin{subfigure}{0.49\columnwidth}
        \includegraphics[width=\linewidth]{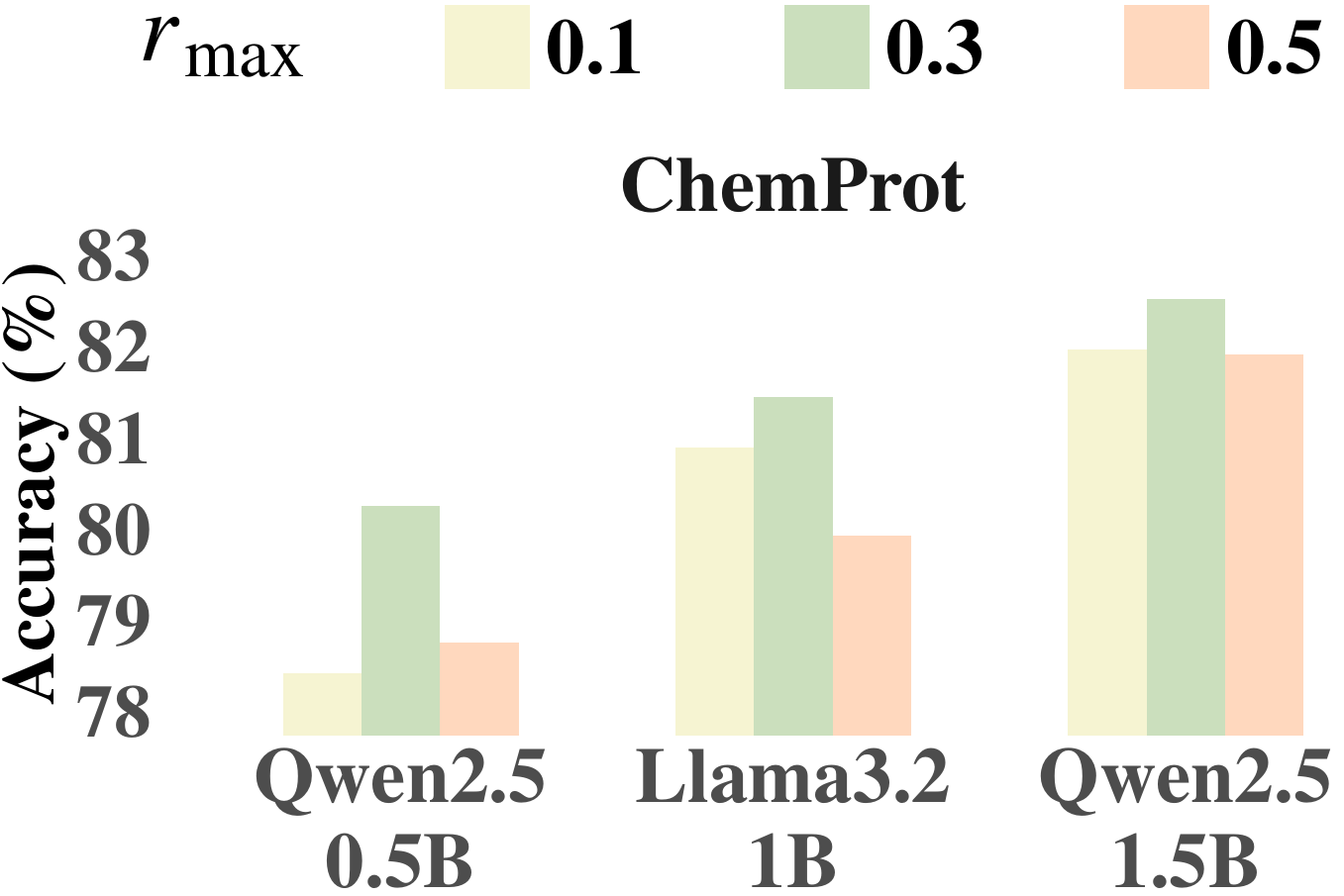}
        \caption{Influence of \( r_{\text{max}} \).}
        \label{fig:r_max}
    \end{subfigure}
    \hfill
    \begin{subfigure}{0.49\columnwidth}
        \includegraphics[width=\linewidth]{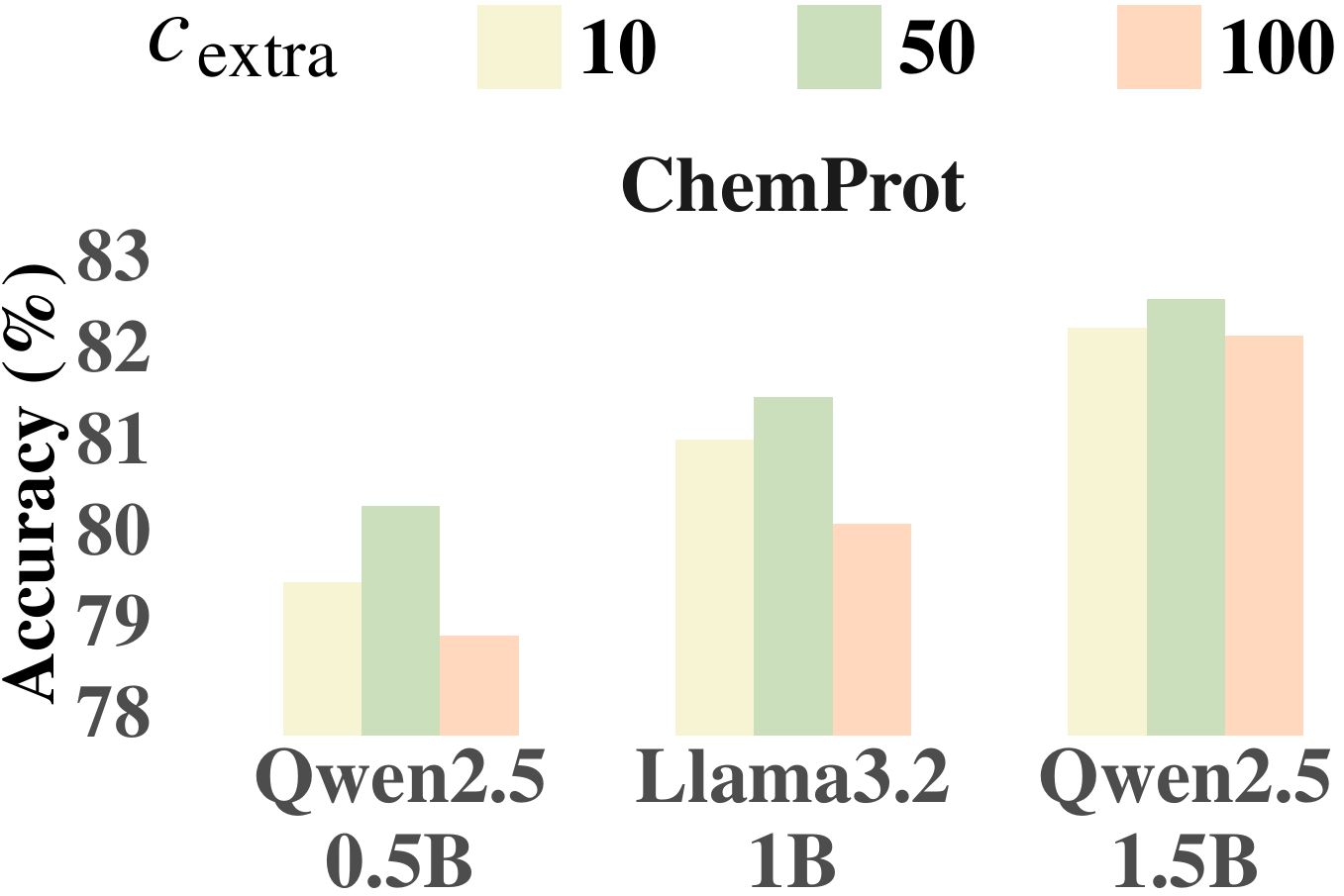}
        \caption{Influence of \( c_{\text{extra}} \).}
        \label{fig:c_extra}
    \end{subfigure}
    \caption{Influence of \( r_{\text{max}} \) and \( c_{\text{extra}} \) on model performance. Figure~\ref{fig:r_max} illustrates the effect of varying \( r_{\text{max}} \) while keeping \( c_{\text{extra}} \) fixed at 50. Figure~\ref{fig:c_extra} shows the effect of varying \( c_{\text{extra}} \) while keeping \( r_{\text{max}} \) fixed at 0.3.}
    \label{fig:combined_r_max_c_extra}
\end{figure}

\noindent \textbf{Max Repetition Ratio}\quad
The choice of \( r_{\text{max}} \) is crucial, as excessively high values increase token repetition, leading to overexposure to redundant context and degraded performance, while low values limit the proportion of overlapping sequences, weakening intended benefits. As shown in Figure~\ref{fig:r_max}, setting \( r_{\text{max}} = 0.3\) achieves a balance between maintaining contextual continuity and preventing excessive redundancy, which is consistent with our theoretical analysis.

\noindent \textbf{Bin Capacity}\quad
Similarly, the choice of \( c_{\text{extra}} \) must be carefully optimized to prevent adverse effects. A large value leads to excessive token truncation, while a very small value provides minimal difference with no extra bin size. Notably, \( c_{\text{extra}} \) should be adjusted in accordance with the sequence length \( L_{\text{seq}} \). The results presented here are based on \( L_{\text{seq}} = 2048 \). As illustrated in Figure~\ref{fig:r_max}, setting \( c_{\text{extra}} = 50 \) achieves the best trade-off between preserving essential context and effectively utilizing the additional capacity.

\subsection{Case Study}
\label{subsec:CaseStudy}
To further demonstrate the importance of contextual continuity, we construct a controlled toy setting. A synthetic event description—unknown to the model—was injected into a corpus of news articles, and \( L_{\text{seq}} \) was set to 32.

We prompt the model with ``\texttt{The Future of Bioengineering Forum will be held on}'' and compare completions from models trained with BFD and our method.

\begin{table}[t]
\centering
\scalebox{0.8}{
\begin{tabular}{lcc}
\toprule
\textbf{Method} & \textbf{Date Accuracy} & \textbf{Location Accuracy} \\
\midrule
\midrule
BFD & 0/5 & 2/5 \\
SP & \textbf{3/5} & \textbf{5/5} \\
\bottomrule
\end{tabular}}
\caption{Generation results of the case study. Full outputs are provided in Appendix~\ref{sec:appendix-case-study}.}
\label{tab:case-study}
\end{table}

The model trained with SP correctly recovers critical information (e.g., date and location), while BFD leads to hallucinated completions. Full generations and discussion are provided in Appendix~\ref{sec:appendix-case-study}.

\subsection{Empirical Analysis}
\label{subsec:EmpiricalAnalysis}
In this section, we conduct extensive experiments to validate the theoretical analysis presented in Section~\ref{subsec:TheoreticalAnalysis} and further investigate key aspects of our method. Unless otherwise specified, all datasets are derived from academic articles in PubMed (\citealp{cohan-etal-2018-discourse}), with detailed data distribution in Appendix~\ref{sec:dataset_length_distribution}. We use the LLaMA-3.2 tokenizer and set the sequence length \( L_{\text{seq}} \) to 2048.

According to Equation~(6), when \( r_{\text{max}} \) is set to 0.3 and \( T_k \) follows the distribution given in Table~\ref{tab:dataset_distribution_PubMed}, the expected number of sequences subject to the sliding window technique, \( N_{sw} \), is 6716. Empirically, we observe that \( N_{sw} \) is 6988, covering approximately 62\% of all texts. The slight discrepancy between theoretical and observed values arises because the real data distribution within each length interval is not perfectly uniform. 

Regarding tokens in short chunks, the theoretical estimate for the total number of short chunks in stage two, denoted as \( N_{\text{token\_short}} \), is 2.6M, while the actual count obtained from experiments is 2.2M, constituting approximately 3.4\% of all tokens. This indicates that the first stage of Seamless Packing plays a significant role, as a substantial portion of tokens is processed during this stage.

Another key consideration is the trade-off between dropping and padding. The empirical observation is that in low-data or short-sequence scenarios, total dropping tokens in SP has small difference with total padding tokens in BFD. However, in high-data or long-sequence scenarios, padding tokens are much more than dropping tokens. Detailed analysis is presented in Appendix~\ref{sec:appendix_empirical_analysis}.

Finally, we analyze the computational efficiency of BFD and FFD. On average, BFD requires 29\% more processing time compared to FFD. While the gap varying across datasets with different distributions, FFD consistently outperforms BFD in efficiency. One possible explanation for this variation is that BFD’s strategy of selecting the most optimally fitting bin introduces additional computational overhead, making FFD the preferred choice for time-sensitive applications. The detailed results are presented in the Appendix~\ref{sec:appendix_FFD_BFD_time}.

\section{Conclusion}
In this study, we introduce Seamless Packing, an innovative methodology for data packing in continual pre-training scenarios. We provide a rigorous theoretical analysis of our method to demonstrate its effect range. Through a comprehensive series of experiments, we have systematically validated the effectiveness of our proposed approach. The empirical results reveal that our sliding window technique significantly enhances context preservation, while the Packing with Dropping strategy demonstrates remarkable efficacy in processing short text segments. These findings underscore the important role of data engineering in the model training paradigm, particularly highlighting that suboptimal data packing methods can substantially limit the utility of even high-quality datasets. We believe that our methodological contributions and empirical insights not only advance the current understanding of data packing in continual learning but also provide valuable perspectives for future research in this domain.

\section*{Limitations}
While our experimental analysis has provided valuable insights into the token dropping and padding dynamics, it is important to note that a comprehensive theoretical framework explaining these phenomena remains to be developed. Besides, our empirical validation, though conducted across various domains, represents a subset of potential application scenarios. The generalizability of our method to other domains, particularly specialized areas such as code generation and analysis, warrants further investigation. Additionally, while our method demonstrates promising results in continual pre-training scenarios, its effectiveness in standard pre-training setting remains unexplored. We have outlined a systematic research agenda to address them in future work.

\section*{Acknowledgments}
The authors would like to thank the anonymous reviewers for their valuable comments. This work was supported by National Natural Science Foundation of China (No. 62076068).

\bibliography{anthology, main}

\clearpage
\appendix

\section{Related Work}
\label{sec:appendix_relatedwork}
\subsection{Continual Pre-training}
Despite the benefits of continual pre-training, it introduces significant challenges, with catastrophic forgetting \citep{Robins1995CatastrophicFR, french1999catastrophic} being the most important. This phenomenon occurs when a model loses previously acquired knowledge while adapting to new data.

To mitigate catastrophic forgetting, various strategies have been proposed. \citet{ke2023continual} combined soft-masks with contrastive learning to preserve prior knowledge while incorporating new information. \citet{winata-etal-2023-overcoming} highlighted the role of learning rate scheduling in balancing the acquisition of new knowledge with the retention of previously learned information. Other studies \citep{lin-etal-2024-mitigating, alexandrov-etal-2024-mitigating} have demonstrated that model merging can effectively alleviate catastrophic forgetting. Additionally, approaches centered on data distribution management \citep{parmar2024reuse, ibrahim2024simple} have been shown to improve performance by ensuring that new training data complements the original pre-training corpus. These methods are orthogonal to our approach, as they primarily focus on model optimization or data distribution rather than the process of data packing, making them compatible and potentially synergistic when combined.

\subsection{Data Packing}
\citet{shi2024incontext} identified a limitation in standard data packing methods, noting that preceding documents in a sequence often provide no predictive signal for subsequent ones. To overcome this, they proposed a document reordering algorithm that groups related documents together, allowing models to better leverage contextual information. 

\section{Theoretical Analysis}
\label{sec:appendix_theoretical_analysis}
\begin{figure}[p] 
    \centering
    \includegraphics[width=0.8\columnwidth]{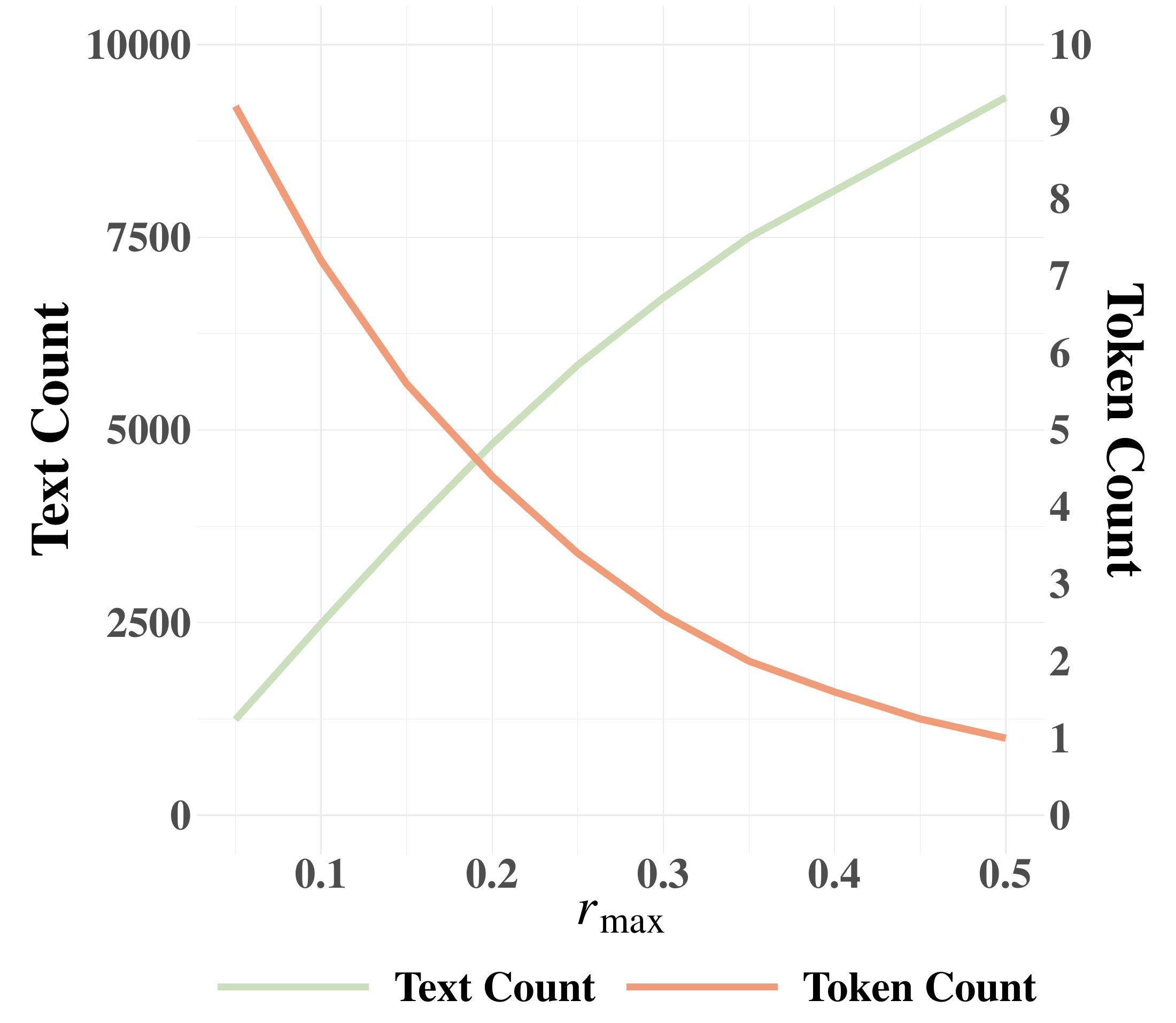}
    \caption{Effect of \( r_{\text{max}} \) on Text Count and Token Count. Text Count: The amount of texts that are able to use sliding window in stage one. Token Count: the total amount of tokens in short chunks in stage two.}
    \label{fig:texttokencount}
\end{figure}

To further illustrate our theoretical analysis, we apply it to the real-world text length distribution and compute the corresponding values for different \( r_{\text{max}} \) settings. Intuitively, the number of texts eligible for sliding window processing in stage one (Text Count) should fall within a reasonable range—if it is too low, the method affects only a small portion of the data, limiting its impact. On the other hand, if it is too high, prior studies suggest that excessive repetition may degrade performance, which has been mentioned before. Similarly, a lower total number of tokens in shorter chunks in stage two (Token Count) is preferable, as we aim for most of the final data to consist of naturally continuous sequences that can fully occupy sequence length, rather than fragmented chunks artificially stitched together. Figure~\ref{fig:texttokencount} presents the result, showing how Text Count and Token Count vary as \( r_{\text{max}} \) increases from 0.05 to 0.5. As expected, Text Count increases with \( r_{\text{max}} \), while Token Count decreases. However, it is also notable that the rate of change in both curves gradually slows down, indicating diminishing returns as \( r_{\text{max}} \) grows. Based on this observation, setting \( r_{\text{max}} \) around 0.3 appears to be a reasonable choice for this dataset, balancing sequence utilization and token repetition.

\section{Dataset Length Distribution}
\label{sec:dataset_length_distribution}
\begin{figure}[p] 
    \centering
    \includegraphics[width=\columnwidth]{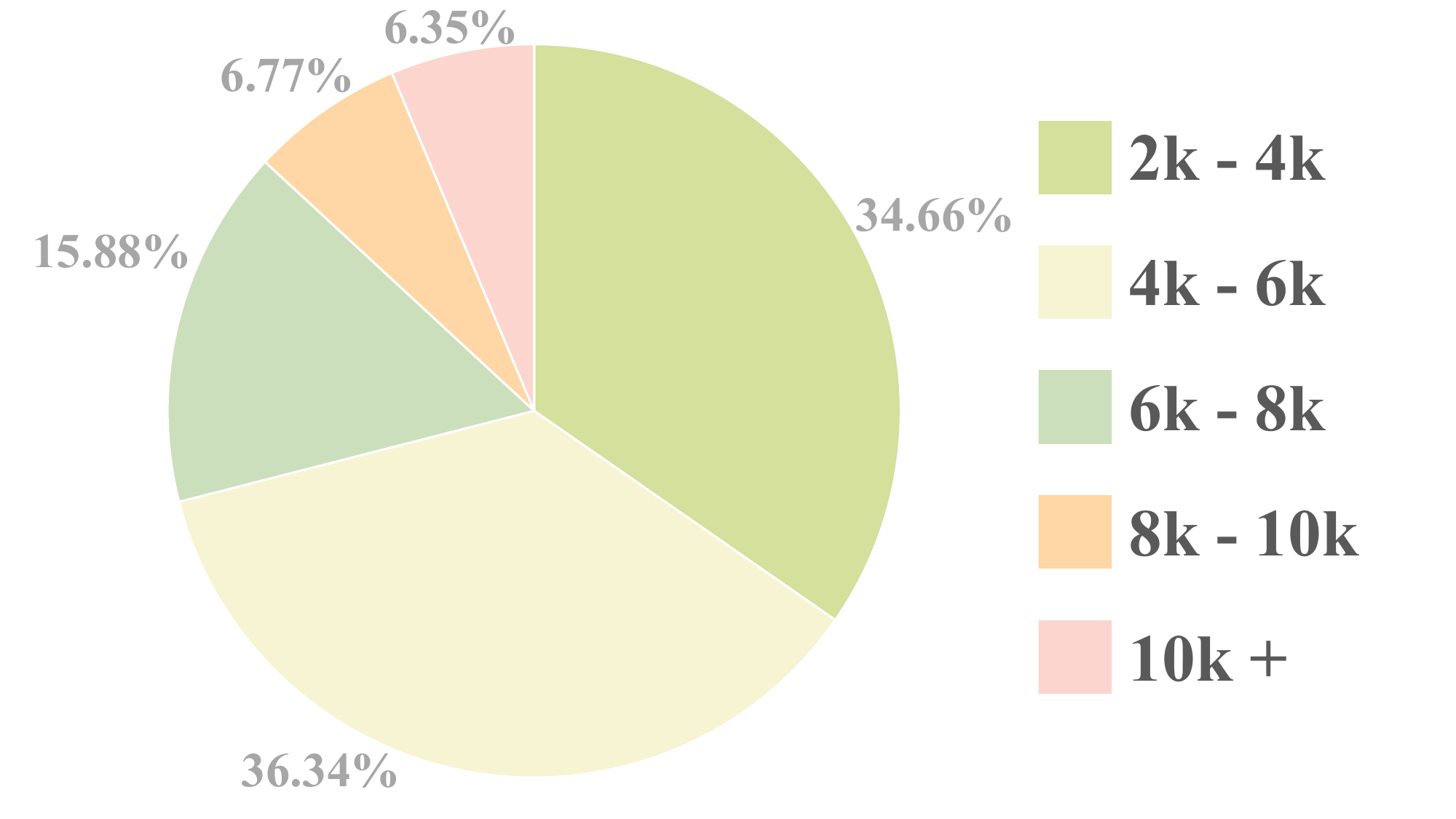}
    \caption{Illustration of text length distribution of the PubMed article Dataset.}
    \label{fig:pie}
\end{figure}

The detailed length distribution of each dataset is presented in Table~\ref{tab:dataset_distribution_PubMed} and Table~\ref{tab:dataset_distribution_combined}. Figure~\ref{fig:pie} offers a more intuitive visualization. While the distributions vary across datasets, they exhibit a common pattern: the majority of texts are relatively short, and the number of texts gradually decreases as the length increases.

\section{Experiment Details}
\label{sec:appendix_experiment_related}

\subsection{Dataset}
The datasets used in Section~\ref{subsec:performance} are derived from BBC News \citep{li2024latesteval}, financial news articles \citep{financearticle}, and academic articles from PubMed \citep{cohan-etal-2018-discourse}, all of which are in English.

For the BBC News and financial news datasets, we set the sequence length \( L_{\text{seq}} \) to 512 and the extra bin capacity \( c_{\text{extra}} \) to 10. We sample both datasets at a ratio of 25\% within each interval, as detailed in Table~\ref{tab:dataset_distribution_combined}. 

For the PubMed dataset, we set \( L_{\text{seq}} \) to 2048 (1024 for GPT-2, which is its maximum sequence length ) and \( c_{\text{extra}} \) to 50. We sample the dataset at a ratio of 17\% within each interval, as specified in Table~\ref{tab:dataset_distribution_PubMed}. For all dataset, we set \( r_{\text{max}} \) to 0.3.

\subsection{Benchmark}
We evaluate our method on multiple benchmark datasets across different domains:

\paragraph{News Domain}  
The BBC News Topic Classification dataset (referred to as BBC News) contains 2,225 articles published on the BBC News website between 2004 and 2005 \citep{greene06icml}. AG News (referred to as AG News) consists of over 1 million news articles \citep{agnews}. The 20 Newsgroups dataset includes approximately 18,000 newsgroup posts spanning 20 topics \citep{LANG1995331}.  

\paragraph{Financial Domain}  
Following \citet{huang-etal-2023-adasent}, we use three benchmarks in the financial domain. The Twitter Financial News dataset is an English-language corpus of finance-related tweets, from which two different benchmarks are derived:  
\begin{itemize}
    \item Financial News Sentiment\footnote{\url{https://huggingface.co/datasets/zeroshot/twitter-financial-news-sentiment}}  
    \item Financial News Topic\footnote{\url{https://huggingface.co/datasets/zeroshot/twitter-financial-news-topic}}  
\end{itemize}  
Additionally, we use the Financial PhraseBank, a sentiment classification dataset containing 4,840 sentences extracted from financial news articles \citep{Malo2014GoodDO}.  

\paragraph{Medical Domain.}  
For medical text classification, we evaluate on two datasets:  
\begin{itemize}
    \item PubMed Text Classification\footnote{\url{https://huggingface.co/datasets/ml4PubMed/PubMed-text-classification-cased}}, a classification dataset derived from PubMed articles.  
    \item ChemProt, from the BioCreative VI Chemical-Protein (ChemProt) interaction dataset, which identifies chemical and protein entities along with their relationships \citep{DBLP:journals/biodb/LiSJSWLDMWL16}.  
\end{itemize}  

For most benchmarks, we down-sample the datasets due to their large size. The actual dataset sizes and label distributions used in our experiments are detailed in Table~\ref{tab:benchmark_detail}.  

\subsection{Baseline}
The baselines we compare our method with include five configurations in total, with details described in the following.

\begin{enumerate}
\setlength{\itemsep}{0pt}
\setlength{\parsep}{0pt}
\setlength{\parskip}{0pt}
    \item \textbf{The original model} (OM). We evaluate downstream tasks using the original model without any continual pre-training, serving as a foundational baseline.
    \item \textbf{Concatenation and Truncation} (CT). This conventional approach processes data by concatenating all texts and splitting them into fixed-length sequences.
    \item \textbf{First-Fit-Decreasing} (FFD). The texts are first segmented into chunks, each with a maximum length of $L_\text{seq}$, where $L_\text{seq}$ denotes sequence length. These chunks are then treated as a bin packing problem, with bin capacity set to $L_\text{seq}$. The First-Fit-Decreasing algorithm is applied, placing each chunk into the first available bin that can accommodate it.
    \item \textbf{Best-Fit-Decreasing} (BFD). Following the same setting as FFD, BFD uses Best-Fit-Decreasing as the packing algorithm, which selects the bin that results in the least remaining space after placement.
    \item \textbf{Modified Best-Fit-Decreasing} (BFD-m). This variant of BFD retains the same packing strategy as BFD but bin capacity is set slightly larger than $L_\text{seq}$, following the second stage of our method Seamless Packing.
\end{enumerate}

\subsection{Continual Pre-training Setup}
We perform continual pre-training on all models with the AdamW optimizer \citep{loshchilov2018decoupled}. We use a learning rate of 5e-5 with a linear learning rate scheduler, and warm up over the first 500 steps. 

For our experiments, models were trained on different GPU configurations based on their parameter sizes:
\begin{itemize}
\setlength{\itemsep}{0pt}
\setlength{\parsep}{0pt}
\setlength{\parskip}{0pt}
    \item Models smaller than 1B parameters were trained on four NVIDIA 2080 Ti GPUs (11GB VRAM each).
    \item Llama-3.2-1B was trained on four NVIDIA 3090 GPUs (24GB VRAM each).
    \item Models larger than 1B parameters were trained on two NVIDIA L20 GPUs (48GB VRAM each).
\end{itemize}

Each continual pre-training run was conducted for 3 epochs, with the total GPU compute time per run ranging from 30 to 100 GPU hours, depending on the model size and dataset.

\subsection{Tuning Setup}
We perform full parameter fine-tuning and LoRA tuning on all models with the AdamW optimizer. We set the learning rate between 1e-5 and 5e-5 for full parameter fine-tuning, depending on the downstream task, and use a fixed learning rate of 1e-4 for LoRA tuning. Both configurations employ a linear learning rate scheduler.

\section{Statistical Significance Analysis}
\label{sec:appendix_stats_significance}
To assess the robustness of our performance improvements, we conducted paired significance tests under our main experimental setup using Qwen-2.5-1.5B. We compared: (1) CT vs. SP , and (2) BFD vs. SP. For each pair, we applied both paired t-tests and Wilcoxon signed-rank tests over five independent runs. Detailed results are shown in table~\ref{tab:stats_test}.

For CT vs. SP, 6 out of 8 tasks yielded t-test p-values below 0.1, and 7 out of 8 tasks yielded Wilcoxon p-values at or below 0.1. For BFD vs. SP, 5 out of 8 (t-test) and 6 out of 8 (Wilcoxon) were below this threshold. This suggests that the improvements from Seamless Packing are statistically meaningful overall.

\paragraph{Note on Wilcoxon Test} The Wilcoxon signed-rank test has a granularity limit with five runs, where the smallest possible p-value is 0.0625. Thus, p-values below 0.05 cannot be achieved under this setup.

\section{Optimizer and Learning Rate Scheduler}
\label{sec:appendix_lr}
To examine the impact of optimizer and scheduler choices on continual pre-training, we conducted additional experiments under the Mixed-Domain Pre-training setting (as in Section~\ref{sec:generalization-analysis}). We evaluated three configurations: (1) SGD with a cosine scheduler, (2) AdamW with a cosine scheduler, and (3) AdamW with a linear scheduler (default). Results are shown in table~\ref{tab:lr}.

Overall, AdamW consistently outperforms SGD, likely due to its adaptive learning rates and better regularization. While both the Linear and Cosine schedulers achieve comparable results, the Linear scheduler consistently yields slightly better performance.

\section{Hyperparameters}
\label{sec:appendix_hyperparameter}
The detailed experimental results for hyperparameter selection are presented in Table~\ref{tab:maxrepetitionratio} and Table~\ref{tab:extrabincapacity}. Notably, a moderate choice of $r_{\text{max}}$ and $c_{\text{extra}}$ yields the highest performance.

\section{Case Study}
\label{sec:appendix-case-study}

We show the full generation outputs for the toy case study introduced in Section~\ref{subsec:CaseStudy}. 

This artificial event is constructed to ensure that the model has no prior knowledge of it, allowing for a clean evaluation of information retention. The inserted content is as follows:

\begin{quote}
According to the company, the \textbf{Future of Bioengineering Forum} explores groundbreaking developments in synthetic biology, including lab-grown organs and genetic enhancements. On \textbf{February 28th, 2028}, at the \textbf{BioTech Conference Center}, experts will discuss both technological progress and ethical dilemmas surrounding human genetic modifications.
\end{quote}

Each model generates five completions, and the outputs are reported below. For clarity, we highlight dates and locations in the responses: \textcolor{softgreen}{green} indicates correct information, while \textcolor{softred}{red} denotes hallucinated or incorrect content.

\paragraph{BFD (Baseline)}
\begin{quote}
The Future of Bioengineering Forum will be held on \textcolor{softred}{October 3rd, 2029}, at the \textcolor{softgreen}{BioTech Conference Center}. \\
The Future of Bioengineering Forum will be held on \textcolor{softred}{February 9th, 2035}, the Department for the Arts and the Department for the Sciences have announced. \\
The Future of Bioengineering Forum will be held on \textcolor{softred}{February 9th, 2037}, to discuss the progress made in synthetic biology. \\
The Future of Bioengineering Forum will be held on \textcolor{softred}{December 8th, 2029}, at the \textcolor{softgreen}{BioTech Conference Center}. \\
The Future of Bioengineering Forum will be held on \textcolor{softred}{December 3rd, 2029}, at the \textcolor{softred}{Bioengineering Conference Center}.
\end{quote}

\paragraph{Seamless Packing (Ours)}
\begin{quote}
The Future of Bioengineering Forum will be held on \textcolor{softgreen}{February 28th,} \textcolor{softred}{2020}, at the \textcolor{softgreen}{BioTech Conference Center}. \\
The Future of Bioengineering Forum will be held on \textcolor{softgreen}{February 28th, 2028}, at the \textcolor{softgreen}{BioTech Conference Center}. \\
The Future of Bioengineering Forum will be held on \textcolor{softgreen}{February 28th, 2028}, at the \textcolor{softgreen}{BioTech Conference Center}. \\
The Future of Bioengineering Forum will be held on \textcolor{softgreen}{February 28th,} \textcolor{softred}{2023}, at the \textcolor{softgreen}{BioTech Conference Center}. \\
The Future of Bioengineering Forum will be held on \textcolor{softgreen}{February 28th, 2028}, at the \textcolor{softgreen}{BioTech Conference Center}.
\end{quote}

\paragraph{Analysis} The completions produced by the BFD-trained model consistently hallucinate event dates and, in some cases, invent irrelevant details. In contrast, the SP-trained model reliably reconstructs the correct date and location, despite some minor inconsistencies in the year (likely due to tokenization artifacts in number handling, such as ``2028'' being split as ``202 + 8'' by the tokenizer). Importantly, the month and day remain correct across all SP completions.

This case study demonstrates that by better preserving contextual continuity, Seamless Packing effectively reduces hallucination and improves factual consistency in downstream tasks.

\section{Empirical Analysis}
\subsection{Dropping and Padding}
\label{sec:appendix_empirical_analysis}
We further analyze the trade-off between dropping and padding. On the PubMed dataset, our method discards 51K tokens, while BFD introduces 75K padding tokens, resulting in a relatively small difference. However, for datasets with significantly different length distributions, such as BBC News, the number of discarded tokens and padding tokens are 7K and 140K, respectively. A comprehensive mathematical proof of this phenomenon is challenging due to the complexity of the underlying distributions, but we provide an intuitive explanation below.

The differences in padding and token dropping behaviors can be better understood through the lens of the FFD algorithm. In low-data or short-sequence scenarios, the limited availability of short texts makes it difficult to efficiently fill partially packed bins, leading to increased padding as the remaining space in most bins cannot be fully utilized. Conversely, in high-data or long-sequence scenarios, the abundance of short text fragments enables more efficient bin packing, thereby reducing padding.

A similar effect is observed in the token dropping strategy. When bins are nearly full, the discarded portion closely aligns with the extra bucket capacity \( c \). In high-data regimes, where most bins are densely packed, the total number of dropped tokens stabilizes around \( c \) per bin, making its impact comparable to that of padding. This explains why dropping tokens is more advantageous than padding in sparse data scenarios but converges to similar efficiency levels in large-scale settings.

\subsection{Time consumption of FFD and BFD}
\label{sec:appendix_FFD_BFD_time}
To further investigate the computational efficiency of First-Fit-Decreasing (FFD) and Best-Fit-Decreasing (BFD), we provide a dataset-wise comparison of their execution times in Table~\ref{tab:FFD_BFD_time}. The results indicate that BFD consistently requires more processing time than FFD, with an average overhead of 29\%.

Examining the dataset-specific results, we observe that the relative efficiency gap between FFD and BFD varies depending on the dataset. For instance, on the BBC News dataset, BFD is approximately 17.4\% slower than FFD, whereas in the Financial Article dataset, the slowdown increases to 27.6\%. The most pronounced difference is found in the PubMed Article dataset, where BFD takes 41.3\% longer than FFD. This suggests that while BFD generally incurs a higher computational cost, the extent of its inefficiency is influenced by the underlying distribution and structure of the data.

One possible explanation for this variation is that BFD’s strategy of selecting the most optimally fitting bin introduces additional computational overhead, especially when dealing with a higher number of fragmented sequences. In contrast, FFD operates with a simpler heuristic, leading to faster assignments at the cost of potentially less compact packing.

Despite these variations, FFD consistently demonstrates superior efficiency across all datasets, reinforcing its suitability for time-sensitive applications. While BFD may still be preferable in scenarios where minimizing fragmentation is the primary concern, its increased computational burden should be carefully considered when scaling to large datasets.

\begin{table}[p]
    \centering
    \begin{tabular}{ccc}
    \toprule
        \textbf{Dataset} & \textbf{FFD} (s) & \textbf{BFD} (s) \\
    \midrule
    \midrule
        BBC News & 2.24 & 2.63 \\
        Financial Article & 2.83 & 3.61 \\
        PubMed Article & 1.26 & 1.78 \\
    \bottomrule
    \end{tabular}
    \caption{Execution time (in seconds) of First-Fit-Decreasing and Best-Fit-Decreasing on datasets.}
    \label{tab:FFD_BFD_time}
\end{table}

\begin{table}[p]
\centering
\resizebox{0.9\columnwidth}{!}{
\begin{tabular}{ccccc}
\toprule
\multirow{2}{*}{\textbf{Model}} & \multirow{2}{*}{\textbf{Metric/Task}} & \multicolumn{3}{c}{\textbf{$r_{\text{max}}$}} \\
\cmidrule(lr){3-5}
& & 0.1 & 0.3 & 0.5 \\
\midrule
\midrule
\multirow{2}{*}{\shortstack{Qwen2.5 \\ \\ 0.5B}} 
& PubMed Class & 84.83 & \textbf{86.12} & 84.96 \\
& ChemProt & 78.43 & \textbf{80.26} & 78.76 \\
\midrule
\multirow{2}{*}{\shortstack{Llama3.2 \\ \\ 1B}} 
& PubMed Class & 84.66 & \textbf{85.47} & 84.72 \\
& ChemProt & 80.90 & \textbf{81.45} & 79.94 \\
\midrule
\multirow{2}{*}{\shortstack{Qwen2.5 \\ \\ 1.5B}} 
& PubMed Class & 85.30 & \textbf{86.56} & 85.75 \\
& ChemProt & 81.97 & \textbf{82.53} & 81.92 \\
\bottomrule
\end{tabular}}
\caption{Performance on downstream tasks using different max repetition ratio, {$c_{\text{extra}}$} is fixed to 50}
\label{tab:maxrepetitionratio}
\end{table}

\begin{table}[p]
\centering
\resizebox{0.9\columnwidth}{!}{
\begin{tabular}{ccccc}
\toprule
\multirow{2}{*}{\textbf{Model}} & \multirow{2}{*}{\textbf{Metric/Task}} & \multicolumn{3}{c}{\textbf{$c_{\text{extra}}$}} \\
\cmidrule(lr){3-5}
& & 10 & 50 & 100 \\
\midrule
\midrule
\multirow{2}{*}{\shortstack{Qwen2.5 \\ \\ 0.5B}} 
& PubMed Class & 84.89 & \textbf{86.12} & 85.17 \\
& ChemProt & 79.42 & \textbf{80.26} & 78.84 \\
\midrule
\multirow{2}{*}{\shortstack{Llama3.2 \\ \\ 1B}} 
& PubMed Class & 85.10 & \textbf{85.47} & 84.79 \\
& ChemProt & 80.99 & \textbf{81.45} & 80.06 \\
\midrule
\multirow{2}{*}{\shortstack{Qwen2.5 \\ \\ 1.5B}} 
& PubMed Class & 85.54 & \textbf{86.56} & 85.74 \\
& ChemProt & 82.21 & \textbf{82.53} & 82.12 \\
\bottomrule
\end{tabular}}
\caption{Performance on downstream tasks using different extra bin capacity, {$r_{\text{max}}$} is fixed to 0.3}
\label{tab:extrabincapacity}
\end{table}

\begin{table}[p]
\centering
\resizebox{\columnwidth}{!}{
\begin{tabular}{lcccc}
\toprule
\multirow{2}{*}{\textbf{Task}} & \multicolumn{2}{c}{\textbf{CT vs SP}} & \multicolumn{2}{c}{\textbf{BFD vs SP}} \\
\cmidrule(lr){2-3} \cmidrule(lr){4-5}
 & \textbf{t-test p} & \textbf{Wilcoxon p} & \textbf{t-test p} & \textbf{Wilcoxon p} \\
\midrule
\midrule
BBC-news       & 0.1151 & 0.0679 & 0.2563 & 0.3125 \\
AG-news        & 0.0297 & 0.0625 & 0.0014 & 0.0625 \\
20 Newsgroup   & 0.3154 & 0.1797 & 0.1217 & 0.0625 \\
Fin Topic      & 0.0084 & 0.0625 & 0.0026 & 0.0625 \\
Fin Sentiment  & 0.0777 & 0.0656 & 0.7552 & 0.4652 \\
Fin Phrasebank & 0.0260 & 0.0625 & 0.0341 & 0.0625 \\
PubMed Class   & 0.0832 & 0.0679 & 0.0450 & 0.0625 \\
ChemProt       & 0.0978 & 0.0679 & 0.0076 & 0.0625 \\
\bottomrule
\end{tabular}}
\caption{P-values from paired t-tests and Wilcoxon signed-rank tests comparing SP with CT and BFD.}
\label{tab:stats_test}
\end{table}

\begin{table}[p]
\centering
\resizebox{\columnwidth}{!}{
\begin{tabular}{cccc}
\toprule
\textbf{Lower Bound} & \textbf{Upper Bound} & \textbf{Number of} & \textbf{Percentage} \\
(k) & (k) & \textbf{Samples} & (\%) \\
\midrule
\midrule
2 & 4 & 3906 & 34.66 \\
4 & 6 & 4095 & 36.34 \\
6 & 8 & 1789 & 15.88 \\
8 & 10 & 763 & 6.77 \\
10 & 12 & 355 & 3.15 \\
12 & 14 & 150 & 1.33 \\
14 & 16 & 73 & 0.65 \\
16 & 18 & 47 & 0.42 \\
18 & $\infty$ & 90 & 0.80 \\
\bottomrule
\end{tabular}}
\caption{Text length distribution of the PubMed article Dataset.}
\label{tab:dataset_distribution_PubMed}
\end{table}

\begin{table*}[p]
\centering
\resizebox{0.9\textwidth}{!}{
\begin{tabular}{cccccc}
\toprule
\multirow{2}{*}{\textbf{Lower Bound}} & \multirow{2}{*}{\textbf{Upper Bound}} & \multicolumn{2}{c}{\phantom{XXX}\textbf{BBC News Dataset}\phantom{XXX}} & \multicolumn{2}{c}{\textbf{Financial Article Dataset}} \\
\cmidrule{3-6}
 (k) & (k) & \phantom{XXX}\textbf{Samples} & \textbf{(\%)} & \phantom{X}\textbf{Samples} & \textbf{(\%)} \\
\midrule
\midrule
0.5 & 1 & \phantom{XXX}3533 & 29.53 & \phantom{X}7790 & 50.17 \\
1 & 1.5 & \phantom{XXX}6556 & 54.79 & \phantom{X}4161 & 26.80 \\
1.5 & 2 & \phantom{XXX}1323 & 11.06 & \phantom{X}1012 & 6.52 \\
2 & 2.5 & \phantom{XXX}330 & 2.76 & \phantom{X}397 & 2.56 \\
2.5 & 3 & \phantom{XXX}97 & 0.81 & \phantom{X}274 & 1.76 \\
3 & 3.5 & \phantom{XXX}49 & 0.41 & \phantom{X}234 & 1.51 \\
3.5 & 4 & \phantom{XXX}27 & 0.23 & \phantom{X}212 & 1.37 \\
4 & 4.5 & \phantom{XXX}22 & 0.18 & \phantom{X}195 & 1.26 \\
4.5 & $\infty$ & \phantom{XXX}29 & 0.24 & \phantom{X}1252 & 8.06 \\
\bottomrule
\end{tabular}}
\caption{Text length distribution of the BBC News and Financial Article datasets.}
\label{tab:dataset_distribution_combined}
\end{table*}

\begin{table*}[p]
\centering
\resizebox{0.8\textwidth}{!}{
\begin{tabular}{ccccc}
\toprule
\textbf{Domain} & \textbf{Benchmark} & \textbf{Train Set Size} & \textbf{Test Set Size} & \textbf{Num of Labels} \\
\midrule
\midrule
\multirow{3}{*}{News} & BBC News & 1225 & 1000 & 5 \\
 & AG News & 1200 & 760 & 4 \\
 & 20 Newsgroup & 1123 & 743 & 20 \\
\midrule
\multirow{3}{*}{Finance} & Financial Topic & 1689 & 402 & 20 \\
& Financial Sentiment & 1907 & 477 & 3 \\
& Financial Phrasebank & 1500 & 1000 & 3 \\
\midrule
\multirow{2}{*}{Medical} & PubMed Class & 1545 & 588 & 5 \\
& ChemProt & 1662 & 689 & 13 \\
\midrule
\multirow{3}{*}{GLUE} & MNLI & 1176 & 980 & 3 \\
& QNLI & 1046 & 1092 & 2 \\
& RTE & 1244 & 277 & 2 \\
\midrule
French & XNLI & 1569 & 1002 & 3 \\
\bottomrule
\end{tabular}}
\caption{Details of benchmarks.}
\label{tab:benchmark_detail}
\end{table*}

\begin{table*}[p]
\centering
\resizebox{0.9\textwidth}{!}{
\begin{tabular}{lcccccc}
\toprule
& News & Finance & PubMed & Mixed & Redpajama & French \\
\midrule
\midrule
Tokens & 15,239,955 & 27,401,686 & 63,260,896 & 50,455,586 & 11,415,242 & 13,531,672 \\ 
\bottomrule
\end{tabular}}
\caption{Token count of training datasets.}
\label{tab:lr}
\end{table*}

\begin{table*}[p]
\centering
\resizebox{0.7\textwidth}{!}{
\begin{tabular}{lccc}
\toprule
\textbf{Task} & \textbf{SGD + Cosine} & \textbf{AdamW + Cosine} & \textbf{AdamW + Linear} \\
\midrule
\midrule
20 Newsgroup   & 69.20 & 69.31 & \textbf{69.80} \\
Fin Phrasebank & 83.04 & 84.38 & \textbf{84.58} \\
ChemProt       & 81.94 & 82.23 & \textbf{82.58} \\
\bottomrule
\end{tabular}}
\caption{Continual pre-training performance under different optimizer and scheduler configurations.}
\label{tab:lr}
\end{table*}

\end{document}